\documentclass[11pt,a4paper]{article}
\usepackage[hyperref]{emnlp2020}
\usepackage{times}
\usepackage{url}
\usepackage{latexsym}
\usepackage{graphicx}
\usepackage{amssymb}
\usepackage{amsmath}
\usepackage{multirow}

\usepackage{microtype}

\aclfinalcopy 
\def\aclpaperid{177} 

\title{Don't Parse, Insert: Multilingual Semantic Parsing with Insertion Based Decoding}

\author{Qile Zhu\textsuperscript{1}\thanks{~~Work done while interning at Amazon Alexa} , Haidar Khan\textsuperscript{2}, Saleh Soltan\textsuperscript{2}, Stephen Rawls\textsuperscript{2}, and Wael Hamza\textsuperscript{2}
	\\
	\textsuperscript{1}University of Florida, \textsuperscript{2}Amazon Alexa AI \\
	\tt {valder@ufl.edu} \\ \tt {\{khhaida,ssoltan,sterawls,waelhamz\}@amazon.com}\\ }

\date{}

\begin{document}
	\maketitle
	\begin{abstract}
		
		Semantic parsing is one of the key components of natural language understanding systems. 
		A successful parse transforms an input utterance to an action that is easily understood by the system. 
		Many algorithms have been proposed to solve this problem, from conventional rule-based or statistical slot-filling 
		systems to shift-reduce based neural parsers. For complex parsing tasks, the state-of-the-art method is based on autoregressive sequence to sequence models to generate the parse directly. 
		This model is slow at inference time, generating parses in $O(n)$ 
		decoding steps ($n$ is the length of the target sequence). In addition,
		we demonstrate that this method performs poorly in zero-shot cross-lingual transfer learning settings.  
		 In this paper, we propose a non-autoregressive parser which is based on the
		  insertion transformer to overcome these two issues. Our approach 1) speeds up decoding by 3x while
		   outperforming the autoregressive model and 2) significantly improves cross-lingual transfer 
		   in the low-resource setting by 37\% compared to autoregressive baseline. We test our approach on three
			well-known monolingual datasets: ATIS, SNIPS and TOP. For cross lingual semantic parsing, 
			we use the MultiATIS++ and the multilingual TOP datasets.
	\end{abstract}

	\section{Introduction}
	\label{intro}
	 Given a query, a semantic parsing module identifies not only the \textit{intent} (play music, book a flight) 
	 of the query but also extracts necessary \textit{slots} (entities) that further refines the action to perform 
	 (which song to play? Where or when to go?). A traditional rule-based or slot-filling system classifies a query 
	 with one intent and tags each input token~\cite{mesnil2013investigation}. However, supporting more complex queries 
	 that are composed of multiple intents and nested slots is a challenging problem~\cite{gupta2018semantic}. 
	 Gupta et al.~\shortcite{gupta2018semantic} and Einolghozati et al.~\shortcite{einolghozati2019improving} 
	 propose to use a Shift-Reduce parser based on Recurrent Neural Network for these complex queries. 
	 Recently, Rongali et al.~\shortcite{rongali2020don} propose directly generating the parse as a formatted 
	 sequence and design a unified model based on sequence to sequence generation and pointer networks. 
	 Their approach formulates the tagging problem into a generation task in which the target 
	  is constructed by combining all the necessary intents and slots in a flat sequence with no restriction
	   on the semantic parse schema. 
	 
	 A relatively unexplored direction is the cross-lingual transfer problem~\cite{duong2017multilingual,susanto2017neural}, 
	 where the parsing system is trained in a high-resource language and transfered directly to a low-resource language (zero-shot). 
	
	The state-of-the-art model leverages the autoregressive decoder such as Transformer~\cite{vaswani2017attention} and Long-Short-Term Memory (LSTM)~\cite{hochreiter1997long} to generate the target sequence (representing the parse) from left to right. The left to right autoregressive generation constraint has two drawbacks: 1) generating a parse takes $O(n)$ decoding time, where $n$ is the length of the target sequence. This is further exacerbated when paired with standard search algorithms such as beam search. 2) In the cross-lingual setting, autoregressive parsers have difficulty transferring between languages. 
	
	A recent direction in machine translation and natural language generation to speed up sequence to sequence 
	models is non-autoregressive decoding~\cite{stern2019insertion,gu2017non,gu2019levenshtein}. Since the parsing task in the sequence to sequence framework only 
	requires inserting tags rather than generating the whole sequence, an insertion based parser is both faster and more natural for language transfer than an autoregressive parser.
	
	In this paper, we leverage insertion based sequence to sequence models for the semantic parsing problem that require only O(log(n)) decoding time to generate a parse. 
	We enhance the insertion transformer~\cite{stern2019insertion} with the pointer mechanism, since the entities in the source sequence are ensured to appear in the target sequence. Our non-autoregressive based model can also boost the performance on the zero-shot and few-shot cross-lingual setting, in which the model is trained on a high-resource language and tested on low-resource languages. We also introduce a copy source mechanism for the decoder to further improve the cross lingual transfer performance. In this way, the pointer embedding will be replaced by the corresponding outputs from the encoder.
	 We test our proposed model on several well known datasets, TOP~\cite{gupta2018semantic}, 
	 ATIS~\cite{price1990evaluation}, SNIPS~\cite{coucke2018snips}, MultiATIS++~\cite{xu2020end} and multilingual TOP~\cite{xiamultilingual}.
	
	In summary, the main contributions of our work include:
	\begin{itemize}
		\item To our knowledge, we are the first to apply the non-autoregressive framework to the semantic parsing task. Experiments show that our approach can reduce the decoding steps by 66.7\%. By starting generation with the whole source sequence, we can further reduce the number of decoding steps by 82.4\%.
		\item We achieve new state-of-the-art Exact Match (EM) scores on ATIS (89.14), SNIPS (91.00) and TOP (86.74, single model) datasets.
		\item We introduce a copy encoder outputs mechanism and achieve a significant improvement compared to the autoregressive decoder and sequence labeling on the zero-shot and few-shot setting 
		in cross lingual transfer semantic parsing. Our approach surpasses the autoregressive baseline by 9 EM points on average over both simple (MultiATIS++) and complex (multilingual TOP) queries and matches the performance of the sequence labeling baseline on MultiATIS++.
	\end{itemize}

	\begin{figure*}
		\centering
		\includegraphics[width=\textwidth]{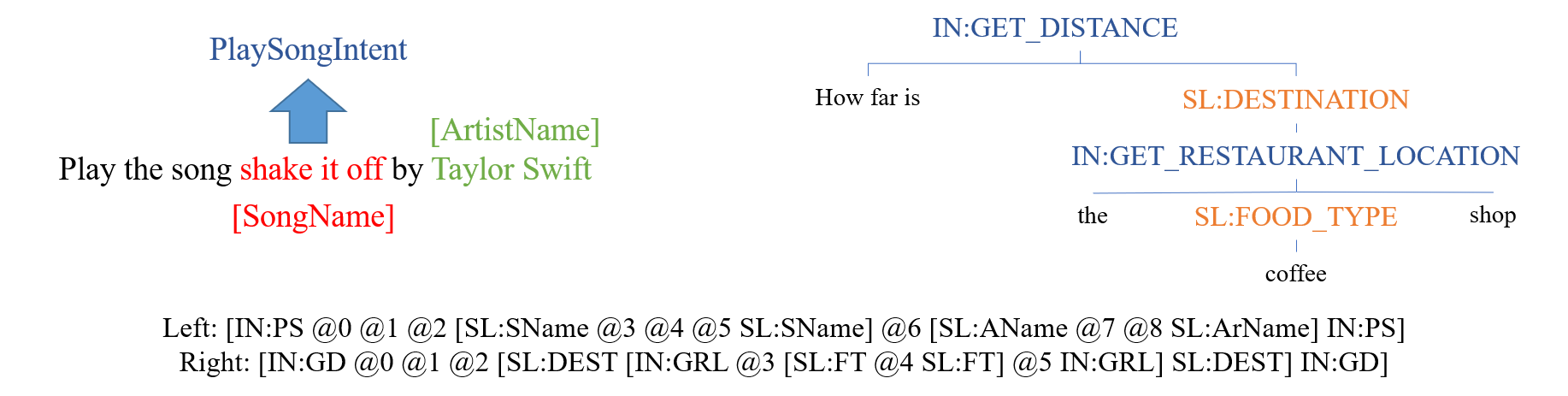}
		\caption{Example of a simple query (left) and complex query (right). The complex query contains multiple intents and nested slots and can be represented as a tree structure. The two queries are represented as formatted sequences that are treated as the target sequence in the parsing task. IN is the intent, SL is the slot. Source tokens that appear in the target sequence are replaced by pointers with the form \textit{@n} where \textit{n} denotes its location in the source sequence. For complex queries, we can build the parse from top to bottom and left to right.}
		\label{fig:example}
	\end{figure*}
	
	\section{Background}
	\label{sec:background}
	In this section, we introduce the sequence generation via insertion operations and the pretrained models we leverage in our work.

	\subsection{Sequence Generation Via Insertion}
	We begin by briefly describing sequence generation via insertion, for a more complete description see ~\cite{stern2019insertion}. 
	
	Let $x_1, x_2 , ..., x_m$ be the source sequence with length $m$ and $y_1, y_2, ..., y_n$ denotes the target sequence with length $n$. We define the generated sequence $h_t$ at decoding step $t$. In the autoregressive setting, $h_t = y_{1,2,...,t-1}$. In insertion based decoding, $h_t$ is a subsequence of the target sequence $y$ that preserves order. For example, if the final sequence $y=[A,B,C,D,E]$, then $h_t=[B,E]$ is a valid intermediate subsequence while $h_t=[C,A]$ is an invalid intermediate subsequence.
	
	During decoding step $t+1$, we insert tokens into $h_t$. In the previous example, there are three available insertion slots: before token $B$, between $B$ and $E$ and after $E$. We always add special tokens such as $bos$ (begin of the sequence) and $eos$ (end of the sequence) to the subsequences. The number of available insertion slots will be $T-1$ where $T$ is the length of $h_t$ including $bos$ and $eos$. If we insert one token in all available slots, multiple tokens can be generated in one time step. 
	
	In order to predict the token to insert in a slot, we form the representation for each insertion slot by pooling the representations of adjacent tokens. We have $T-1$ slots for a sequence with length $T$. Let $r \in \mathbb{R}^{T\times d}$, where $T$ is the sequence length and $d$ denotes the hidden size of the transformer decoder layer. All slots $s \in \mathbb{R}^{(T-1) \times d}$ can be computed as:
	
	\begin{align}
	s = concat(r[1:], r[:-1]) \cdot W_{s},
	\end{align}
	where $r[1:]$ is the entire sequence representation excluding the first token, $r[:-1]$ is the entire sequence representation excluding the last token and $W_s \in \mathbb{R}^{2d \times d}$ is a trainable projection matrix. We apply $softmax$ to the slot representations to obtain the token probabilities to insert at each slot.
	
	\subsection{Pretrained Models}
	Pretrained language models \cite{devlin2018bert,liu2019roberta,lan2019albert,dong2019unified,peters2018deep} have sparked significant progress in a wide variety of natural language processing tasks. The basic idea of these models is to leverage the knowledge from large-scale corpora by using a language modeling objective to learn a representation for tokens and sentences. For downstream tasks, the learned representations are fine-tuned for the task. This improvement is even more significant when the downstream task has few labeled examples. 
	
	We also follow this trend, and use the Transformer~\cite{vaswani2017attention} based pretrained language model. We use the RoBERTa base~\cite{liu2019roberta} (we refer to this model as RoBERTa) as our query encoder to fairly compare with the previous method. This model has the same architecture as BERT base~\cite{devlin2018bert} with several modifications during pretraining. It uses a dynamic masking scheme and removes the next sentence prediction task. RoBERTa is also trained with longer sentences and larger batch sizes with more training samples. For the multilingual zero-shot and few-shot semantic parsing task, 
	we use XLM-R~\cite{conneau2019unsupervised} and multilingual BERT~\cite{devlin2018bert} which are trained on text for more than 100 languages.
	
	\section{Methodology}
	
	In this section, we introduce our non-autoregressive sequence to sequence model for the semantic parsing problem.
	
	\subsection{Query Formulation}
	To train a sequence to sequence model, we prepare a source sequence and a target sequence. For the task of semantic parsing, the source sequence is the query in natural language. We construct the target sequence following Rongali et al.~\shortcite{rongali2020don} and Einolghozati et al. \shortcite{einolghozati2019improving}. Tokens in the source sequence that are present in the target sequence are replaced with the special pointer token \textit{ptr-n}, where \textit{n} is the position of that token in the source sequence. By using pointers in the decoder, we can drastically reduce the vocabulary size. We follow previous work and use symmetrical tags for \textit{intents} and \textit{slots}. Fig.~\ref{fig:example} shows two examples, a simple query and a complex query with the corresponding target sequences. This formulation is also able to express other tagging problems like named entity recognition (NER). 
	
	\subsection{Insertion Transformer}
	
	We use the insertion transformer~\cite{stern2019insertion} as the base framework for the decoder. The insertion transformer is a modification of the original transformer decoder architecture~\cite{vaswani2017attention}. The original transformer decoder predicts the next token based on the previously generated sequence while the insertion transformer can predict tokens for all the available slots. In this setup, tokens in the decoder side can attend to the entire sequence instead of only their left side. This means we remove the causal self-attention mask in the original decoder.
		
	\subsubsection{Pointer Network with Copy}
	\textbf{Pointer Network:} In the normal sequence to sequence model, target tokens are generated by feeding the final representations (decoder hidden states) through a feed-forward layer and applying a softmax function over the whole target vocabulary. This is slow when the vocabulary size is large~\cite{yang2017breaking}.  In parsing, the entities in the source sequence will always appear in the target sequence. We can leverage the pointer mechanism~\cite{vinyals2015pointer} to reduce the target vocabulary size by dividing the vocabulary into two types: \textit{tokens} that are the parsing symbols like intent and slot names, and \textit{pointers} to words in the source sequence.  
	
	Since we have two kinds of target tokens, we use two slightly different ways to obtain unnormalized probabilities for each type. For the tokens in the tagging vocabulary, we feed the hidden states generated by the insertion transformer and slot pooling to a dense layer to produce the logits of size \textit{V} (tagging vocabulary). The tagging vocabulary contains only the parse symbols like intents and slots together with several special tokens such as \textit{bos}, \textit{eos}, the padding and unknown token. For the pointers, we compute the scaled dot product attention scores between the slot representation and the encoder output. The attention scores will be computed as 
	
	\begin{align}
	a(Q,K) = \frac{QK^{T}}{\sqrt{h}},
	\end{align}
	
	where query ($Q$) is the slot representation, the encoder outputs would be the key ($K$) and $h$ is the hidden size of the query. Since the hidden size of encoder and decoder may be different, we also do a projection of query and key to the same dimension with two dense layers. Notice that the length of attention scores follows the length of the source sequence. Concatenating the attention scores with size $n$ and the logits for the tagging vocabulary (\textit{V}), we get the unnormalized distribution over $V+n$ tokens. We apply the $softmax$ function to obtain the final distribution over these tokens. 
	\\
	\noindent
	\textbf{Copy Mechanism:} Rongali et al.~\shortcite{rongali2020don} use a set of special embeddings to represent pointer tokens. This is a problem because the pointer embedding cannot encode semantic information since it points to different words across examples. Instead, we reuse the encoder output that the pointer token points to. Without copying, the special pointer embedding would learn a special position based representation for the source language that is hard to transfer to other languages. 
	
	\subsection{Training and Loss}
	\label{sect:loss}
	
	Training the insertion decoder requires sampling source and target sequences from the training data. 
	We randomly sample valid subsequences from the target sequence to mimic intermediate insertion steps. 
	 We first sample a length $k \in [0,n]$ for the subsequence, where $n$ is the length of the target sequence (here $n$ excludes the $bos$ and $eos$ tokens). We select $k$ tokens from the target sequence and maintain the original ordering. This sampling helps the model learn to insert tokens from the initial generation state as well as intermediate generation.
	
	The insertion transformer can do parallel decoding since we can insert tokens in all available insertion slots. However, for each insertion slot, there may be multiple candidate tokens that can be inserted. For example, given a target sequence $[A,B,C,D,E]$ and a valid subsequence $[A,E]$, the candidates for the slot between token $A$ and $E$ are $B,C,D$. We use the two different weighting schemes proposed in Stern et al.~\shortcite{stern2019insertion}: uniform weights and balanced binary tree weights. 
	\\
	\noindent
	\textbf{Binary Tree Weights:} The motivation for applying binary tree weighting is to make the decoding time nearly O(log(n)). 
	Consider the example of sequence $A,B,C,D,E$ again, the desired order of generation would 
	be $[bos,eos] \rightarrow [bos,C,eos] \rightarrow [bos,A,C,E,eos] \rightarrow [bos,A,B,C,D,E,eos]$. 
	To achieve this goal, we weight the candidates according to their positions. For the sequence above, candidates in the span of $[bos, eos]$ are
	$A,B,C,D,E$. We assign token $C$ the highest weight, then lower weights for $B,D$ and the lowest weights for $A,E$.
	
	Given a sampled subsequence with length $k+1$, we have $k$ insertion slots at location $l=(0,1,...,k-1)$. Let $c_{l_0},...c_{l_i}$ be the candidates for one location $l$. We can define a distance function $d_j$ for each token $j$ in the candidates of $l$:
	\begin{align}
	d_{l}(j) = |j - \frac{i}{2}|,
	\end{align}
	where $i$ is the number of candidates in the location $l$. We then use the negative distance to compute the softmax based weighting~\cite{rusu2015policy,norouzi2016reward}:
	\begin{align}
	w_l(j) = \frac{exp(-d_l(j) / \tau)}{\sum_{m=0}^{i}exp(-d_l(m)/\tau)}.
	\end{align}
	Where $\tau$ is the temperature hyperparameter which allows us to control the sharpness of the weight distribution. 
	\\
	\noindent
	\textbf{Uniform Weights:} Instead of encouraging the model to follow a tree structure generation order, we can also treat the candidates equally. This performs better than the binary tree weights when we input the whole source sequence to the decoder as the initial sequence. In this case, we only need to insert the tagging tokens; the number of candidates is not as large as from scratch ($[bos, eos]$). This uniform weighting can be easily done by taking $\tau \rightarrow \infty$. 
	\\
	\noindent
	\textbf{Loss Function:}The autoregressive sequence to sequence model uses the negative log-likelihood loss since in each decoding step, there is only one ground-truth label. However, in our approach, we have multiple candidates for each insertion slot. Therefore, we  use the KL-divergence between the predicted token distribution and the ground truth distribution. Then the loss for insertion slot $l$ is:
	\begin{align}
	L_{slot}(x,h_t,l) = D_{KL}((p_l|(x,h_t)) || g_l),
	\end{align}
	where $p_l$ is the distribution output by the decoder and $g_l$ is the target distribution where we set the probability to 0 for tokens that are not candidates. Note that the ground truth distribution depends on the weighting scheme for generation.
	
	Finally, we have the complete loss averaged over all the insertion slots:
	\begin{align}
	L(x,h_t) = \frac{1}{k+1}\sum_{l=0}^{k}L_{slot}(x,h_t,l)
	\end{align}
	
	\subsection{Termination Strategy}
	Terminating generation for insertion based decoding is not as straightforward as autoregressive decoding, 
	which only needs the no-insertion token to be predicted. Insertion decoding requires a similar mechanism for 
	every insertion slot. When computing the slot-loss above, if there are no candidates for the slot we set the 
	ground truth label as the no-insertion token. At inference time, we can stop decoding when all available slots 
	predict the no-insertion token. However, there is a problem when combining the sampling method and this termination 
	strategy. The no-insertion token is more frequent compared with other tokens. The same situation is also encountered 
	in~\cite{stern2019insertion}. This is solved by adding a penalty hyperparameter to control the sequence length 
	generated by the decoder. The hyperparameter is simply a scalar subtracted from the log probability of the 
	no-insertion token for each insertion slot during inference. By doing this, we set a threshold for the difference
	 between the no-insertion token and the second-best choice.
	
		\begin{table*}[]
		\centering
		\begin{tabular}{ccccccc}
			\hline
			& \multicolumn{2}{c}{TOP}                                                                & \multicolumn{2}{c}{ATIS}        & \multicolumn{2}{c}{SNIPS}       \\
			\multirow{-2}{*}{Method}  & EM                                                                    & IC             & EM             & IC             & EM             & IC             \\ \hline
			Joint BiRNN~\cite{hakkani2016multi}              & -                                                                     & -              & 80.70          & 92.60          & 73.20          & 96.90          \\
			Attention BiRNN~\cite{liu2016attention}           & -                                                                     & -              & 78.90          & 91.10          & 74.10          & 96.70          \\
			Slot Gated Full Attention~\cite{goo2018slot} & -                                                                     & -              & 82.20          & 93.60          & 75.50          & 97.00          \\
			CapsuleNlU~\cite{zhang2018joint}                & -                                                                     & -              & 83.40          & 95.00          & 80.90          & 97.30          \\
			SR(\textit{S})+ELMO+SVMRank~\cite{gupta2018semantic}           & 83.93                                                       & -              & -              & -              & -              & -              \\
			SR(\textit{E})+ELMO+SVMRank~\cite{gupta2018semantic}           & \textbf{87.25}                                                        & -              & -              & -              & -              & -              \\
			AR-S2S-PTR (paper)~\cite{rongali2020don}        & 86.67                                                                 & 98.13          & 87.12          & \textbf{97.42} & 87.14          & 98.00          \\
			AR-S2S-PTR (reproduce)~\cite{rongali2020don}     &  85.67 & 98.17          & 88.91          & 97.09          & 90.71          & \textbf{98.43}          \\ \hline
			IT-S2S-PTR ($\tau$ = 1)               & \textbf{86.74}                                                                 & 98.47 & \textbf{89.14} & 97.31          & \textbf{91.00} & \textbf{98.43} \\ 
			IT-S2S-PTR (input-src, uniform)               & 85.41                                                                 & \textbf{98.71} & - & -          & - &- \\ \hline
		\end{tabular}
		\caption{Exact Match and Intent Classification scores for on the test set. Input-src means the initial input of the decoder is the whole source sequence.
		For the shift reduce parsing models, \textit{E} denotes the ensemble model and \textit{S} is the single model. }
		\label{tb:result}
	\end{table*}
	\begin{table*}[]
		\centering
		\begin{tabular}{ccccccccccc} \\ \hline
			\multirow{2}{*}{Model}  & \multirow{2}{*}{Avg. steps} & \multicolumn{9}{c}{\# tokens generated per step}                      \\
			&                                     & 1    & 2    & 3    & 4    & 5    & 6    & 7    & 8   & 9   \\ \hline
			AR-S2S-PTR            & 17.7                                & 1    & 1    & 1    & 1    & 1    & 1    & 1    & 1   & 1   \\
			IT-S2S-PTR            & 5.9                                 & 1.0  & 2.0  & 3.96 & 6.66 & 6.24 & 3.17 & 1.6  & 1.4 & 1.2 \\
			IT-S2S-PTR(input-src) & \textbf{3.1}                        & 4.99 & 2.92 & 1.37 & 1.00 & 0.54 & 0.27 & 0.25 & 1.0 & 1.0 \\ \hline
		\end{tabular}
		\caption{Decoding statistics on the TOP dataset. Average target sequence length of TOP is 17.7 tokens, we see that the insertion based parser can fully utilize binary tree decoding. "input-src" means we set the whole source sequence as the initial decoder state. }
		\label{tb:step}
	\end{table*}
	\section{Experiments}
	\label{sec:exp}
	In this section, we introduce the datasets and baseline models we experiment with. Then we report the results of monolingual experiments and cross lingual transfer learning experiments. 
	
	\subsection{Datasets}
	\label{ssec:data}
	
	\subsubsection{SNIPS}
	The SNIPS dataset~\cite{coucke2018snips} is a public dataset aimed to improve the semantic parsing models. It contains seven different intents: SearchCreativeWork,
	GetWeather, BookRestaurant, PlayMusic, AddToPlaylist, RateBook, and SearchScreeningEvent. For each intent, there are about 2000 training samples and 100 test samples. The SNIPS dataset consists of only simple queries. 
	
	\subsubsection{ATIS}
	The Airline Travel Information System (ATIS)~\cite{price1990evaluation} dataset was originally collected in the early 90s. The utterances are transcribed from the audio recordings of flight reservation calls. Similar to SNIPS, it consists of only simple queries. ATIS contains seventeen different intents. However, nearly 70\% of the queries are the FLIGHT intent. 
	
	Recently, a multilingual version of ATIS called MultiATIS++ is introduced by Xu~\shortcite{xu2020end}. It is an extension of the Multilingual ATIS~\cite{upadhyay2018almost}. Besides the original three languages (English, Hindi and Turkish), MultiATIS++ adds six new languages including Spanish, German, Chinese, Japanese, Portuguese and French annotated by human experts and consists of a total of 37,084 training samples and 7,859 test samples. We exclude Turkish in our experiments as the test set size is limited.
	
	\subsubsection{TOP}
	Since ATIS and SNIPS contain only simple queries, the Facebook Task Oriented Parsing (TOP) 
	dataset~\cite{gupta2018semantic} was introduced for complex hierarchical and nested queries 
	that are more challenging. The dataset contains around 45,000 annotated queries with 25 intents and 36 slots. 
	They further split them into training (31,000), validation (5,000) and test (9,000). 
	As shown in Fig.~\ref{fig:example}, the nested slots make it harder to parse using a simple sequence tagging model. 
	We also do experiments on multilingual TOP~\cite{xiamultilingual} with Italian and Japanese data. 
	In this dataset, the training and validation set is machine translated while the test set is annotated by human experts.
	
	\subsection{Baseline Models}
	\label{ssec:baseline}
	
	\textbf{Monolingual Baselines:} For monolingual experiments, we select the algorithms reported in Rongali et al.~\shortcite{rongali2020don} as baselines for ATIS and SNIPS. Two of them leverage the power of RNNs: with attention~\cite{liu2016attention} and without attention~\cite{hakkani2016multi}. Another model works completely with attention~\cite{goo2018slot}. A Capsule Networks based model is also included~\cite{zhang2018joint}. Finally, we compare with the autoregressive sequence to sequence with pointer model which is most recent~\cite{rongali2020don}. Simple tagging based models cannot easily handle the complex queries in the TOP dataset. For the TOP dataset, we compare with two previous models, a shift reduce parsing model~\cite{gupta2018semantic} and the autoregressive sequence to sequence model~\cite{rongali2020don}. For all monolingual experiments, we use RoBERTa as our pretrained encoder~\cite{liu2019roberta}. \\
	\textbf{Cross lingual Baselines:} For multilingual experiments (zero-shot and few-shot), 
	we use a sequence labelling model based on multilingual BERT and an autoregressive sequence to sequence model~\cite{rongali2020don} as our baseline. To make fair comparasion, we also use the copy source mechanism in the AR model. For sequence labeling, instead of using F1 score, we also use the exact match (EM) which requires all intents and slots are labeled correctly by the model.
	
	\subsection{Results}

	\begin{table*}[]
		\centering
		\begin{tabular}{cccccccccc} \\ \hline
			  & en    & es    & pt    & de    & fr    & hi    & zh    & ja$^{*}$    & avg   \\ \hline
		IT-S2S-PTR    & \textbf{87.23} & \textbf{50.06} & \textbf{39.30} & \textbf{39.46} & \textbf{46.78} & 11.42 & \textbf{28.72} & 12.60 & 32.69 \\
		AR-S2S-PTR   & 86.83 & 40.72 & 33.38 & 34.00 & 17.22 & 7.45  & 23.74 & 10.04 & 23.77 \\
		mBERT & 86.33 & 48.46 & 38.56 & 39.12 & 42.98 & \textbf{15.22} & 21.89 & \textbf{23.29} & 32.78 \\ \hline
		\end{tabular}
		\caption{Zero-shot cross lingual EM scores by our approach (IT), autoregressive baseline (AR) and sequence labeling baseline (mBERT). Results are averaged over four random seeds. For our approach, we initialize the decoder with source sequences. $^*$ indicates that the data format for the language is not consistent with the S2S model tokenizer.}
		\label{tb:atisit}
	\end{table*}

	\subsubsection{Model Configuration}
	We use the pretrained RoBERTa and mBERT as the encoder for our model. For the decoder side, we use 4 layers with 12 heads transformer decoder. The hidden size of the decoder is the same as the embedding size of the pretrained encoder. For optimization, we use Adam~\cite{kingma2014adam} with $\beta_1=0.9$ and $\beta_2=0.98$, paired with the Noam learning rate (initialized with 0.15) scheduler~\cite{vaswani2017attention} with 500 warmup steps. For cross-lingual experiments, we freeze the encoder's embedding layer.

	\subsubsection{Monolingual Results}

	\begin{table}[]
		\centering
		\begin{tabular}{cccc} \\ \hline
									   & en    & it    & ja      \\ \hline
		IT-S2S-PTR                     & 84.61 & 50.07 & 3.64  \\
		AR-S2S-PTR & \textbf{85.4} & 41.06 & 0.64  \\ \hline
		\end{tabular}
		\caption{Zero-shot EM scores on multilingual TOP dataset. Model is trained on English only.}
		\label{tb:topzeroshot}
	\end{table}

	\begin{table*}[]
		\centering
		\begin{tabular}{ccccc|cccc} \\ \hline
			& \multicolumn{4}{c}{IT-S2S-PTR} & \multicolumn{4}{c}{AR-S2S-PTR} \\ \hline
			\# samples & 0      & 10    & 50    & 100   & 0      & 10    & 50    & 100   \\ \hline
			it & \textbf{50.07}  & 50.13 & 52.69 & \textbf{56.42} & 41.06  & 42.23 & 44.98 & 46.96 \\
			ja & \textbf{3.64}   & 4.7  & 18.01 & \textbf{18.96} & 0.64   & 1.73  & 10.78 & 18.56 \\ \hline
		\end{tabular}
		\caption{Few-shot EM scores on multilingual TOP dataset with model pretrained on English. Training samples used in few-shot are sampled from the test set and excluded during testing.}
		\label{tb:topfewshot}
	\end{table*}

	We use the exact match (EM) accuracy as the main metric to measure the performance of different models. By using EM, the entire parsing sequence predicted by the model has to match the reference sequence, since it's not easy to apply the F1 score or semantic error rate~\cite{thomson2012n} to complex queries. It's better to use the EM here for both simple and complex queries. We also report the intent classification accuracy for our models.
	
	\noindent
	\textbf{Main Result:} Table~\ref{tb:result} shows the results from monolingual experiments on three datasets: TOP, ATIS and SNIPS. Our insertion transformer with pointer achieves new state-of-the-art performance on ATIS and SNIPS under EM metric. For TOP dataset, our model matches the best performance reported for single models (AR-S2S-PTR) despite being 3x faster. 

	We also experiment with starting generation with the entire source sequence as the initial state of the decoder. 
	The performance degrades slightly in this case, possibly due to a training/inference mismatch in this setting. This degradation is likely due to training the model to generate the entire target sequence but only asking the model to generate tags during inference.
	
	\noindent
	\textbf{Decoding Steps:} Since our approach can do parallel decoding, the number of decoding steps is only O(log(n)). Table~\ref{tb:step} shows the statistics for the average decoding steps for the TOP dataset and the number of generated tokens per step. The insertion transformer with pointer only needs 5.9 steps while the autoregressive needs 17.7, resulting in a 3x speedup with insertion decoding. The decoding steps can be further reduced to 3.1 when we start decoding with the source sequence as the initial sequence for the decoder. Theoretically, a perfect binary tree based insertion model should generate $2^{n-1}$ tokens for the $n_{th}$ decoding step. We can see that our approach can make full use of the parallel decoding during the first three steps, since the average length of TOP's test samples is only 17.7.
	
	\noindent
	\textbf{Weighting Strategy:} We do experiments on both binary tree weighting and uniform weighting for the TOP dataset. We set $\tau \in [0.5, 1.0, 1.5, 2.0]$ and find 1.0 performs best. Binary tree weights are better than uniform in the setting of decoding from scratch. However, uniform performs better when we decode from the whole source sequence.

	\subsubsection{Cross Lingual Transfer Results}
	
	For MultiATIS++, we train on English training data and test on all languages. Table~\ref{tb:atisit} shows the results of our approach compared to the autoregressive and sequence labeling baselines. We find that:
	\begin{itemize}
		\item Our approach outperforms the baseline on most of the languages except Hindi and Japanese. For Japanese, we found inconsistencies in the tokenizer that is the likely cause of the degradation \footnote{Chinese is tokenized at the character level in mBERT, while Katakana/Hiragana are tokenized with whitespace. Data in MultiATIS++ is mixed in these two fashions.}.
		\item The autoregressive baseline performs poorly on cross lingual experiments. For example, it only achieves 17.22 EM on the French test set while the other two systems achieve >40 EM. This highlights the weakness of autoregressive parsers that cannot produce parses directly from the encoded representations of the source sequence.
		\item The order of the sentence in Hindi and Japanese is different from others, this may limit the performance of transfer learning for S2S parsers.
	\end{itemize}
	
	We also test on the multilingual TOP dataset~\cite{xiamultilingual}, which extends 
	the TOP datasets to other languages providing human annotated Italian and Japanese test sets. 
	TOP contains a much larger test set compared to ATIS. 
	Table~\ref{tb:topzeroshot} shows the zero-shot results and Table~\ref{tb:topfewshot} shows the few-shot results. 
	
	In the zero-shot setting, our approach achieves 50.07 EM score for Italian while AR only achieves 41.06. Both models are unable to achieve good performance in the zero-shot setting for Japanese. We speculate on this behavior in the few-shot experiment results
	
	In the few shot setting, we finetune the model in two stages, first on the entire English data and then with 10, 50, 100 training samples from other languages. Our approach outperforms the AR baseline in all few shot settings. For Italian, increasing training samples from 10 to 100 does not result in much gain, since the knowledge from English can readily be transferred to Italian, probably due to the similarity of the languages. To further improve the performance on Italian, the model may need many more training samples. However, for Japanese little knowledge (like word order) can be transferred from English so both models can perform as if training from scratch. There may be two reasons here: 1) the order of a sentence is different from English. 2) the annotated target is aligned with the original words in the multilingual TOP so the order of pointers are mixed. Thus, we see the EM scores improves drastically as the number of training samples increases.
	
	\begin{table}[h]
		\centering
		\begin{tabular}{cl} \\ 
		\multicolumn{1}{l}{} & EM    \\ \hline
		\multicolumn{1}{l}{IT-S2S-PTR} & 86.74   \\ \hline
		$\tau=0.1$                            & 74.84 \\
		$\tau=0.5$                           & 85.47 \\
		$\tau=1.0$                           & \textbf{86.74} \\
		$\tau=1.5$                          & 86.33 \\ \hline
		\multicolumn{1}{l}{no copy}  & 86.09 \\ \hline
		\end{tabular}
		\caption{The ablation study for the $\tau$ parameter and copy source embedding vector vs. no copy in the monolingual setting. Results on the TOP dataset show the importance of copying source embeddings. We also observe that small values of $\tau$ (i.e. weighting the central token for insertion heavily) degrade performance.}		\label{tb:tau}
	\end{table}

	\begin{table}[]
		\centering
		\begin{tabular}{ll} \\ \hline
		Models          & EM    \\ \hline
		IT-S2S-PTR-Best & 50.07 \\ \hline
		- copy          & 47.00 \\
		- input-src     & 42.03 \\ \hline
		AR-S2S-PTR-BEST & 41.06 \\ \hline
		- copy          & 30.87 \\ \hline
		\end{tabular}
		\caption{The ablation study for source embedding copying and starting generation from source tokens in the cross-lingual setting. Results are zero-shot in Italian. For the IT-S2S model, both copying and starting generation with source tokens contribute to zero-shot performance}
		\label{tb:ablation}
	\end{table}

	\subsection{Ablation Study}

	For ablation study, we separate the experiments to monolingual and multilingual as above. For multilingual experiments, we use the Italian from multilingual TOP dataset. 

	From Table~\ref{tb:tau}, we observe that the copy mechanism improves performance in the monolingual setting. 
	For the hyperparameter $\tau$, recall that a higher value for $\tau$ would result in flatter (more uniform) weights 
	for the candidates. $\tau=1.0$ provides the best balance between equally weighting the candidates and weighting the next token to be inserted heavily. However, we find that when initializing the decoder with source sequences, uniform weights performs better than binary tree weights.

	For cross-lingual experiments, we introduce two
	components to improve the performance. Table~\ref{tb:ablation}
	shows that both of them help in the zero-shot transfer setting. From the results, we can observe that initializing the decoder with the
	source sequence plays an important role in zero-shot transfer, which is impossible for the autoregressive based models. The copy mechanism is again
	beneficial for both the sequence to sequence models, improving the performance of even the autoregressive model from 30.87 EM to 41.06 EM in the zero-shot Italian experiment.
		
	\section{Related Work}
	\textbf{Monolingual Semantic Parsing:} The task oriented semantic parsing for intent classification and slot detection is usually achieved by sequence labeling. Normally, the system will first classify the query based on the sentence level semantic and then label each word in the query. Conditional Random Fields (CRFs)~\cite{peters2018deep,lan2019albert,jiao2006semi} is one of the most successful algorithms applied to this task before deep learning dominated the area. Deep learning algorithms boost the performance of semantic parsing, especially using recurrent neural networks~\cite{liu2016attention,hakkani2016multi}. Other architectures are also explored, such as convolutional neural networks~\cite{kim2014convolutional} and capsule networks~\cite{zhang2018joint}. 
	
	\noindent
	\textbf{Cross Lingual Transfer Semantic Parsing:} Multilingual natural language understanding has been studied in a variety of tasks including part-of-speech (POS) tagging~\cite{plank-agic-2018-distant,yarowsky-etal-2001-inducing,tackstrom2013token}, named entity recognition~\cite{zirikly-hagiwara-2015-cross,tsai-etal-2016-cross,xie-etal-2018-neural} and semantic parsing~\cite{xu2020end}. Before the advent of pretrained cross-lingual language models, researchers leveraged the representations learned by multilingual neural machine translation (NMT). Another approach is to use NMT to translate between the source language and the target language. However, it is challenging for the sequence tagging tasks: labels on the source language need to be projected on the translated sentences~\cite{xu2020end}. Pretrained cross-lingual language models~\cite{devlin2018bert,lample2019cross} achieve great success in various multilingual natural language tasks.

	\section{Conclusion}
	
	In this paper, we tackle two shortcomings of the autoregressive sequence to sequence semantic parsing models: 
	1) expensive decoding and 2) poor cross-lingual performance. 
	
	We propose 1) insertion transformer with pointers and 2) a copy mechanism which replaces the pointer embeding with corresponding encoder outputs, to mitigate these two problems.
	Our model can achieve O(log(n)) decoding time with parallel decoding. 
	For the specific task of semantic parsing, we can further reduce the decoding steps 
	by initializing the decoder sequence with the whole source sequence. 
	Our model achieves new state-of-the-art performance on both simple queries (ATIS and SNIPS) and complex queries (TOP).
	In cross-lingual transfer, our approach surpasses the baselines in the zero-shot 
	setting by 9 EM points on average across 9 languages.
	
	

	\bibliographystyle{acl_natbib}
	\bibliography{emnlp2020}

\end{document}